# A Machine Learning-based Framework for Predictive Maintenance of Semiconductor Laser for Optical Communication

Khouloud Abdelli, Helmut Grießer, *Member, IEEE,* Stephan Pachnicke, Senior *Member, IEEE*

*Abstract*—Semiconductor lasers, one of the key components for optical communication systems, have been rapidly evolving to meet the requirements of next generation optical networks with respect to high speed, low power consumption, small form factor etc. However, these demands have brought severe challenges to the semiconductor laser reliability. Therefore, a great deal of attention has been devoted to improving it and thereby ensuring reliable transmission. In this paper, a predictive maintenance framework using machine learning techniques is proposed for real-time heath monitoring and prognosis of semiconductor laser and thus enhancing its reliability. The proposed approach is composed of three stages: i) real-time performance degradation prediction, ii) degradation detection, and iii) remaining useful life (RUL) prediction. First of all, an attention based gated recurrent unit (GRU) model is adopted for real-time prediction of performance degradation. Then, a convolutional autoencoder is used to detect the degradation or abnormal behavior of a laser, given the predicted degradation performance values. Once an abnormal state is detected, a RUL prediction model based on attention-based deep learning is utilized. Afterwards, the estimated RUL is input for decision making and maintenance planning. The proposed framework is validated using experimental data derived from accelerated aging tests conducted for semiconductor tunable lasers. The proposed approach achieves a very good degradation performance prediction capability with a small root mean square error (RMSE) of 0.01, a good anomaly detection accuracy of 94.24% and a better RUL estimation capability compared to the existing ML-based laser RUL prediction models.

## I. Introduction

SEMICONDUCTOR lasers have been widely used as optical communication light sources for high speed data transmission due to their high efficiency, low cost, and compactness. Since its invention in 1962, the performance and the productivity of semiconductor lasers have been extensively improved to meet the demands of next generation high speed optical networks in terms of linewidth, power consumption, cost etc. [1,2]. However, the performance of the laser during operation can be adversely affected by several intrinsic and external factors such as contamination [3,4], facet oxidation [5], threading dislocations in the substrate [6], crystal defects, a high ambient temperature [3], etc. Many of these factors are hard to predict but induce laser degradation and failure, and thereby result in optical network disruption and high maintenance costs. Moreover, the lifetime of the laser device is prone to a wear-out failure mode (i.e., gradual degradation) defined as the usual failure mode of a device operating over its service [7]. The complexity of the laser structure, and the diversity of the factors inducing the degradation make the reliability assessment a challenging issue [3]. Therefore, a great deal of research has been devoted to improving the laser reliability.

The qualification of laser reliability is typically performed with laboratory data, obtained from accelerated life tests conducted under high stress conditions such as high temperatures or high drive current. This speeds up the degradation and thereby shortens the time to failure of the device, otherwise the time required to collect field lifetime data from operational devices can be years [8]. Conventionally, the laser lifetime is estimated by extrapolating a mathematical fit of the laser current or output power over time. However, such a reliability extrapolation is inaccurate and can result in considerable overestimation or underestimation of the actual lifetime of the laser. The laser is considered degraded if the value crosses the threshold, which is determined based on the laser design and specifications. However, the threshold approach is imprecise and leads to a high false alarm rate. Recently, machine learning (ML) concepts achieving higher accuracy and prediction capability have been proposed to improve the laser reliability estimation. Abdelli et al. [9, 10] proposed a federated learning approach for semiconductor laser lifetime prediction, and developed an artificial neural network model for laser mean time to failure (MTTF) prediction given the laser characteristics. However, the degradation trend over time, which impacts the estimation of MTTF, is not taken into consideration as features for the ML model. We also presented

This work has been performed in the framework of the CELTIC-NEXT project AI-NET-PROTECT (Project ID C2019/3-4), and it is partly funded by the German Federal Ministry of Education and Research (FKZ16KIS1279K).

K. Abdelli is with ADVA Optical Networking SE, Fraunhoferstr. 9a, 82152 Munich/Martinsried, Germany, and with Kiel University (CAU), Chair of Communications, Kaiserstr. 2, 24143 Kiel, Germany (e-mail: kabdelli@adva.com).

H. Grießer is with ADVA Optical Networking SE, Fraunhoferstr. 9a, 82152 Munich/Martinsried, Germany (e-mail: HGriesser@adva.com).

S. Pachnicke is with Kiel University (CAU), Chair of Communications, Kaiserstr. 2, 24143 Kiel, Germany (e-mail: stephan.pachnicke@tf.uni-kiel.de).

a long short-term memory (LSTM) model for laser failure modes, trained with synthetic data modelling the different laser degradation types [11], and we proposed a hybrid prognostic model based on convolutional neural networks (CNN) and LSTM for laser remaining useful life prediction (RUL), trained with experimental data [12]. However, due to the limited amount of the data used to train the RUL prediction model, the performance of the model was not good.

In this paper, an ML-based framework for predictive maintenance of a semiconductor laser is proposed for monitoring and performing diagnosis and prognosis of the laser device during operation once deployed in an optical network. The proposed framework is composed of three phases: real-time monitoring, degradation detection and RUL prediction. The proposed approach is trained with synthetic laser reliability data produced by a generative adversarial neural network (GAN) model and validated using experimental data of tunable lasers. Our main contributions can be summarized as follows:

- A predictive maintenance framework using different ML techniques to enhance the reliability of the semiconductor laser during operation, and thereby maximizing the overall equipment effectiveness and minimizing the costs and effectively scheduling the maintenance activities
- An ML model for real-time prediction of the performance degradation, adopting the combination of a gated recurrent unit (GRU) and an attention mechanism.
- A convolutional autoencoder model for laser degradation or abnormal behavior.
- An attention-based deep learning model by making full use of the combination of the statistical features characterizing the degradation trend and the temporal sequential features.
- The proposed framework is validated using experimental data of a tunable laser, and the results demonstrate the effectiveness of the framework by achieving high prediction capability and detection accuracy.
- A GAN model for generating realistic laser reliability data.

The rest of this paper is structured as follows: Section 2 gives some background information about GRU, autoencoder, attention mechanism and GAN. Section 3 presents the proposed framework as well as the different ML models involved in the development of the framework. Section 4 describes the experimental data, the data generation using GAN and the validation of the presented framework. Conclusions are drawn in Section 5.

## II. BACKGROUND

In this section, we briefly describe the theoretical concepts about the machine learning models involved in the development and the validation of the proposed framework.

### A. Gated recurrent unit (GRU)

The GRU, recently proposed by Cho et al. in 2014 to tackle the gradient vanishing problem [13], is an improved version of standard recurrent neural networks (RNNs), used to process sequential data and to capture long-term dependencies. The typical structure of a GRU, shown in Fig. 1, contains two gates, a reset and an update gate, controlling the flow of the information. The update gate regulates the information that flows into the memory, while the reset gate controls the information flowing out the memory. The GRU cell is updated at each time step $t$ by applying the following equations:

$$z_t = \sigma(W_z x_t + W_z h_{t-1} + b_z) \quad (1)$$
$$r_t = \sigma(W_r x_t + W_r h_{t-1} + b_r) \quad (2)$$
$$\widetilde{h_t} = \tanh(W_h x_t + W_h (r_t \circ h_{t-1}) + b_h) \quad (3)$$
$$h_t = z_t \circ h_{t-1} + (1 - z_t) \circ \widetilde{h_t} \quad (4)$$

where $z$ denotes the update gate, $r$ represents the reset gate, $x$ is the input vector, $h$ is the output vector, $W$ and $b$ represent the weight matrix and the bias vector, respectively. $\sigma(\cdot)$ is the gate activation function, $\tanh(\cdot)$ represents the output activation function. "∘" represents the element-wise product operator. Equation (1) represents the update gate, Equation (2) the reset gate, Equation (3) computes a candidate state for the current time step using the parts of the previous hidden state, and Equation (4) shows how the output $h_t$ is calculated.

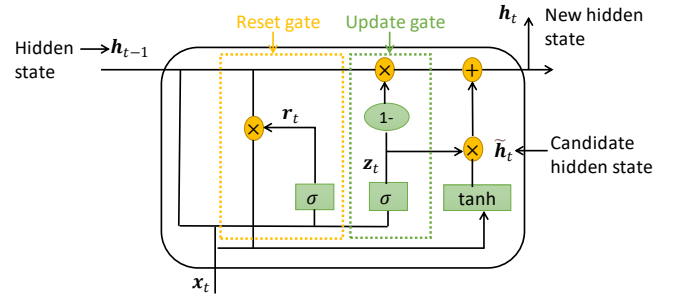

Fig. 1. Structure of the gated recurrent unit (GRU) cell.

### B. Autoencoder

An autoencoder (AE) is a specific type of artificial neural network seeking to learn a compressed representation of an input in an unsupervised manner [14]. An AE is composed of two sub-models, namely the encoder and the decoder. Fig. 2 shows a standard architecture of the AE. The encoder is used to compress an input $x$ into a lower-dimensional encoding (i.e. latent-space representation) $z$ through a non-linear transformation, which is expressed as follows:

$$z = f(Wx + b), \quad (5)$$

where $W$ and $b$ denote the weight matrix and bias vector of the encoder and $f$ represents the activation function of the encoder. The decoder reconstructs the output $\hat{x}$ given the representation $z$ via a nonlinear transformation, which is formulated as follows:

$$\hat{x} = g(W'z + b'), \quad (6)$$

where $W'$ and $b'$ represent the weight matrix and the bias vector of the decoder and $g$ denotes the activation function of the decoder.

The AE is trained by minimizing the reconstruction error between the output $\hat{x}$ and the input $x$, which is the loss function $L(\theta)$, typically the mean square error (MSE), defined as:

$$L(\theta) = \sum \|x - \hat{x}\|^2 \quad (7)$$

where $\theta = \{W, b, W', b'\}$ denotes the set of the parameters to be optimized.

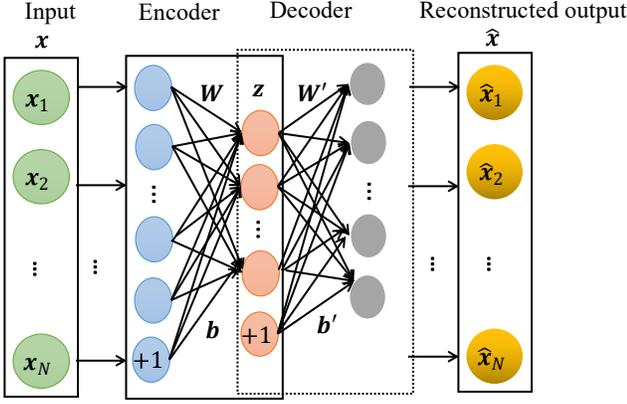

Fig. 2. Structure of a standard autoencoder: the training objective is to minimize the reconstruction error between the output $\hat{x}$ and the input $x$.

*C. Attention mechanism*

Inspired by the human brain that focuses on the distinctive parts rather than processing the entire data, attention mechanisms have been developed to give more 'attention' to the relevant parts of the input while ignoring the others, in order to boost the performance of the deep learning models. The attention mechanism has been applied successfully in many tasks such as speech recognition [15] or machine translation [16].

Let $H = \{h_1, h_2, \ldots h_k\}$ denote the extracted features by a neural network model (i.e., the outputs of the model). The attention mechanism takes $H$ as input and computes an attention score (i.e., weight) $\alpha_i$ for each $h_i$ to decide which features should have more attention. $\alpha_i$ is calculated as follows:

$$e_i = \tanh(W_h\, h_i) \quad (8)$$
$$\alpha_i = \text{softmax}(w^T e_i) \quad (9)$$

where $W_h$, $w$ denote weight matrices. The softmax, a mathematical function that converts a vector of numbers into a vector of probabilities, that sum to one, is used to normalize $\alpha_i$ and to ensure that $\alpha_i \geq 0$, and $\sum_i \alpha_i = 1$. The softmax function can be expressed as follows:

$$\text{softmax}(z)_i = \frac{\exp(z_i)}{\sum_{j=1}^{k} \exp(z_j)} \quad \text{for } i = 1 \ldots k \quad (10)$$

Where $z = (z_1 \ldots z_k) \in \mathbb{R}^k$ denotes the input vector and $z_i$ represents an element of the input $z$.

The different computed weights $\alpha_i$ are aggregated to obtain a weighted feature vector (i.e., attention context vector) $c$, which captures the relevant information to improve the performance of the neural network model. $c$ is computed as follows:

$$c = \sum_i \alpha_i\, h_i \quad (11)$$

*D. Generative adversarial networks (GANs)*

GANs [17] are a type of generative models able to create new content such as an image, text, or audio. The GAN model architecture consists of two sub-models, namely the generator $G$ and the discriminator $D$, which are trained and optimized simultaneously while competing with each other. $G$ is trained to produce realistic samples from a random noise input $z$, to fool $D$, which is trained to distinguish the real samples $x$ from the fake ones made by $G$. The objective function of GAN model ($\mathcal{L}_{GAN}$), whereby $G$ tries to minimize it and $D$ tries to maximize it, can be formulated as follows:

$$\mathcal{L}_{GAN} = \min_G \max_D \; [\, \mathbb{E}_{x \sim p_{data}(x)}[\log(D(x))] + \mathbb{E}_{z \sim p_z(z)}[\log(1 - D(G(z)))]\,] \quad (12)$$

## III. PROPOSED FRAMEWORK

Figure 3 illustrates the proposed predictive maintenance framework for a semiconductor laser. After the deployment of the laser device in an optical network, the current of the laser (i.e. degradation parameter) is monitored periodically under a constant output power. The collected laser current measurements are then stored in a database. Thereafter, the last $k$ monitored current measurements $\{I_{t-k} \ldots I_t\}$, are extracted, preprocessed, and fed to an ML model for real-time prediction of the performance degradation trend. The ML model predicts the next value of the laser current $I_{t+1}$, which is saved in the database. Finally, the sequence of current measurements $\{I_{t-k} \ldots I_{t+1}\}$ is given to an anomaly detection model to identify any degradation or abnormal behavior. If a degradation is detected, a notification is sent to the maintenance planning unit for root cause analysis and an RUL prediction model is triggered to estimate the RUL of the device, which is used then to schedule the maintenance activities.

In the following subsections, the architecture of each model involved in the framework is described.

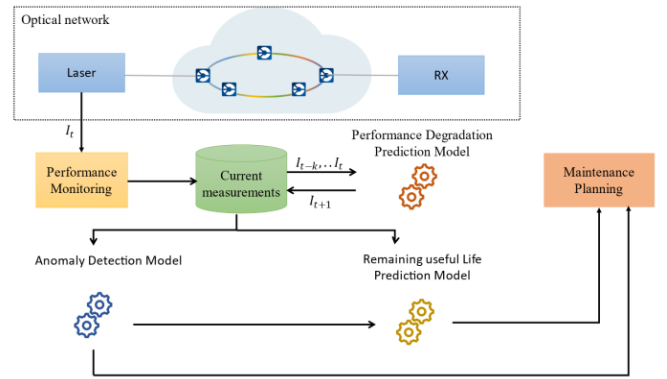

Fig. 3. Flow chart of the proposed framework.

### A. ML based performance degradation prediction model

The proposed ML model for real-time prediction of performance degradation (i.e. increase of laser current) adopts a combination of the GRU and the attention mechanism. The GRU is used to perform a one-step prediction, whereas the attention mechanism helps the model to concentrate more on the relevant features in order to improve the prediction accuracy and to boost the robustness of the model. As shown in Fig. 4, the attention based GRU model takes as input a sequence of length 9 of historical current measurements $[I_{t-8}, I_{t-7},...I_t]$ and predicts the next current measurement $I_{t+1}$. The structure of the proposed model is composed of two GRU layers with 64 and 32 cells, respectively, followed by an attention layer, succeeded by a fully connected layer with no activation function and output $I_{t+1}$. The GRU layers process the sequential input to capture the temporal dependency modelling the degradation trend, and outputs the hidden states $[h_{t-8}, h_{t-7},...h_t]$ (i.e. the learned or extracted features). Then, the attention layer assigns to each extracted feature $h_i$ a weight (i.e. attention score) $\alpha_i$ in order to compute a context vector $c_t$ capturing the relevant information. Afterwards, the weighted attention features are fed to a fully connected layer with output $I_{t+1}$. The model is optimized by minimizing the MSE (i.e. the cost function) between the predicted current value and the true one, by adopting the Adaptive moment estimation (Adam) optimizer.

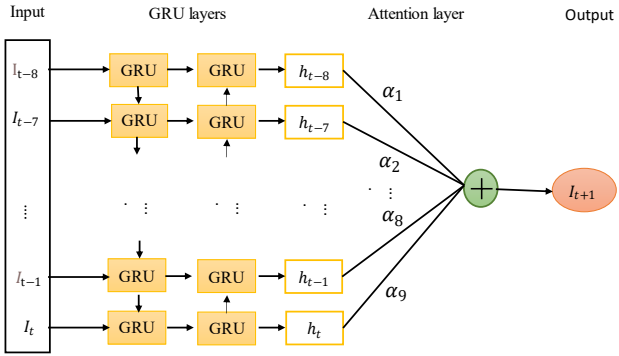

Fig. 4. Architecture of the proposed attention based GRU model for performance degradation prediction.

### B. ML based anomaly detection model

The proposed model for laser anomaly detection is based on a convolutional autoencoder. Please note that the autoencoder is selected as an ML model for the anomaly detection as it is capable of detecting rare or unseen abnormal behavior such as sudden degradation without requiring, neither for the training nor the learning phase, the need to get faulty data representing all types of faults or anomalies that are accurately labeled, which can be prohibitively expensive and cumbersome to obtain. Furthermore, the autoencoder works well for the case of highly unbalanced data (the number of normal samples are higher than the abnormal or faulty samples), which is predominant due to the scarcity of the failures during the system operation of the semiconductor lasers. The architecture of the proposed model is illustrated in Fig. 5. The model contains an encoder and a decoder sub-model with 7 layers. The encoder takes as input a current measurement sequence of length 10. It encodes the input into low dimensional features through a series of 3 convolutional layers containing 32, 16, 32 filters (i.e. kernels) of size 3×1 with a stride (i.e. the step of the convolution operation) of 2, 1, 1, respectively. Then the decoder attempts to reconstruct the input, given the compressed representation output of the encoder. The decoder is inversely symmetric to the encoder part. It consists of 4 transposed convolutional layers used to up-sample the feature maps. The last transposed convolutional layer with one filter of size 3×1 and a stride of 1 is used to generate the output. The Rectified Linear Unit (ReLU) is selected as an activation function for the hidden layers of the model. The cost function is the MSE, which is adjusted by using the Adam optimizer.

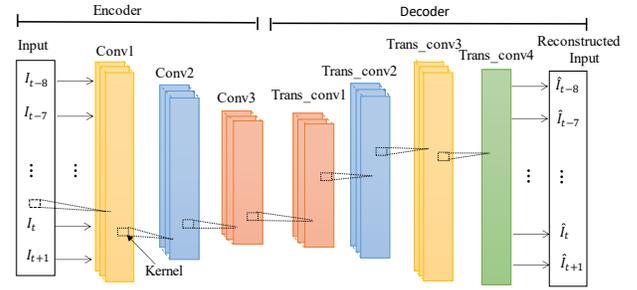

Fig. 5. Structure of the proposed convolutional autoencoder for laser anomaly detection.

Note that the model is trained with normal data modelling the normal state of the laser device in order to learn the distribution characterizing the normal behavior. Once the model is trained, the classification of an instance or observation as anomalous/normal is performed by following the process illustrated in Fig. 6. First, an anomaly score quantifying the distance or the error between the input $I$ and the reconstructed input $\hat{I}$ (the output) is computed. In this study, the mean absolute error (MAE) is selected as an anomaly score. Then, if the calculated anomaly score is higher than a set threshold $\theta$, the instance is classified as "anomalous", else it is assigned as "normal". $\theta$ is a hyperparameter optimized based on the number of true and false positives.

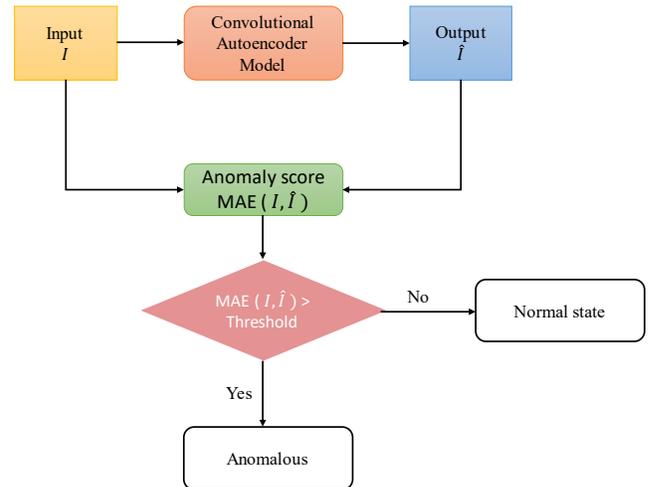

Fig. 6. Flow chart of the process of instance classification as anomalous or normal.

### C. ML based RUL prediction

Figure 7 illustrates the proposed attention-based deep learning model for RUL prediction. The proposed approach makes full use of the fusion of the sequential and temporal features learned by the GRU-attention based layers and the statistical features characterizing the degradation trend, namely the root mean square (RMS), kurtosis ($\beta$) and skewness ($\delta$), in order to improve the prediction accuracy. The sequential input $[I_{t-8}, I_{t-7},…I_{t+1}]$ is fed to 2 GRU layers composed of 64 and 32 cells to learn the representative sequential features, whereas the statistical features are given to a fully connected layer containing 32 neurons. Afterwards, the learned features are transferred to the attention layer to identify the most important features, which are then merged with the features output of the fully connected layer. The fused features are then fed to a fully connected layer with 32 neurons, followed by a dropout layer to avoid overfitting, that finally outputs the RUL. The whole network is simultaneously trained by minimizing the cost function (MSE) with an Adam optimizer.

## IV. VALIDATION OF THE PROPOSED FRAMEWORK

### A. Methodology

To validate the proposed framework, the methodology shown in Fig. 8 is adopted. Firstly, accelerated aging tests under high temperature, to induce the laser degradation and thereby to accelerate the failure of the device, are conducted, whereby the laser current is monitored periodically under constant optical output power. Then, the collected current measurement data is segmented and normalized. Afterwards, the preprocessed data is fed to a GAN model to train it to synthetically generate realistic data that resembles the real data. Once the GAN model is trained, the generator is employed to produce synthetic data which is then used to train the ML model. Afterwards, the trained model is tested with the real data to evaluate the performance of the model.

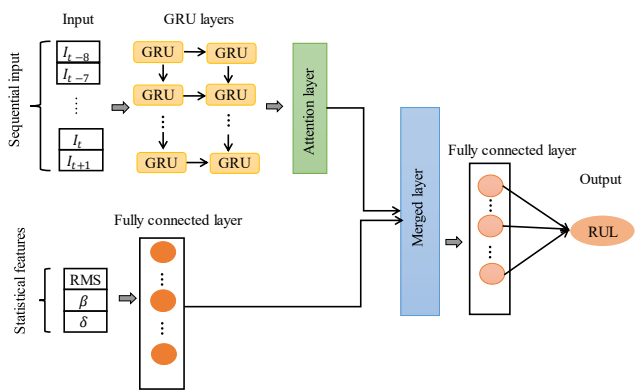

Fig. 7. Structure of the proposed model for laser RUL prediction.

### B. Experimental data

Accelerated aging tests are performed for different tunable laser devices operating at high temperature of 90°C to strongly increase the laser degradation and thereby speed up the device failure. The current is monitored periodically under constant output power at time: 2, 20, 40, 60, 80, 100, 150, 500, 1000, 1500, 2000, and 3000 h. The time to failure of the device $t_f$ is defined as the time at which the current has increased beyond 20% of its initial value. Figure 9 shows the recorded current measurements of the tested devices. It can be observed that some devices failed before the end of the aging test, and that few of the lasers exhibited an abnormal behavior.

In total, the dataset comprised 384 samples incorporating the sequences of monitored current measurements of the tested devices (i.e., 384 semiconductor lasers). We assign to each sample the RUL computed as the difference between $t_f$ and the time $t$ at which the RUL is predicted and the state of the device (normal or anomalous/degraded).

For the training of the GAN model, we consider just the samples of the normal devices as the ML based anomaly detection model is trained just with normal data. Then, the first 10 current values from each considered sample are extracted, whereas the remaining current values are kept for testing the short-term and the long-term prediction capability of the ML based performance prediction model. In total, a dataset of 278 samples is used for GAN model training.

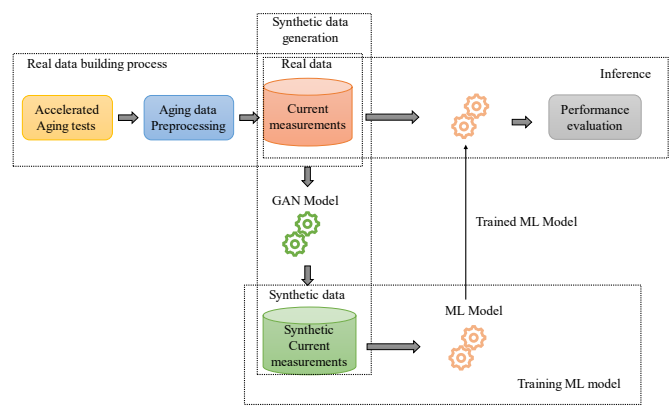

Fig. 8. Methodology for the validation of the proposed framework.

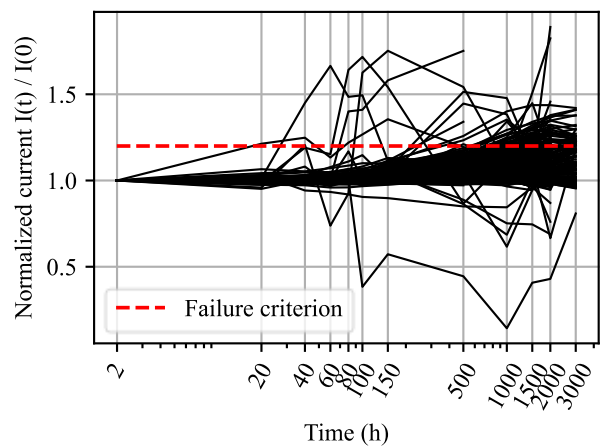

Fig. 9. Recorded current measurements of different laser devices conducted at 90°C.

### C. Synthetic data

The process of the synthetic laser reliability data generation using GAN is illustrated in Fig. 10. The GAN approach is trained with the real data. The generator attempts to produce realistic data from a random noise input, whereas the discriminator is trained to distinguish the fake data generated by the generator and the real data. The generator and the discriminator are updated simultaneously. The training process continues till the generator can generate data samples that the discriminator cannot differentiate from real data. The architecture of the generator is composed of one LSTM layer with 8 cells, followed by 4 convolutional layers containing 32, 16, 16, 1 filters of size 3×1. The discriminator contains 3 convolutional layers with 32 filters of size 3×1. Leaky ReLU, an improved version of ReLU function having a small slope for negative values instead of a flat slope, is set as an activation function for the hidden layers of the generator and the discriminator. The cost function of the GAN model is the binary cross entropy, whereby the generator tries to minimize it and the discriminator tries to maximize it. The optimizer is Adam with a learning rate of 0.001.

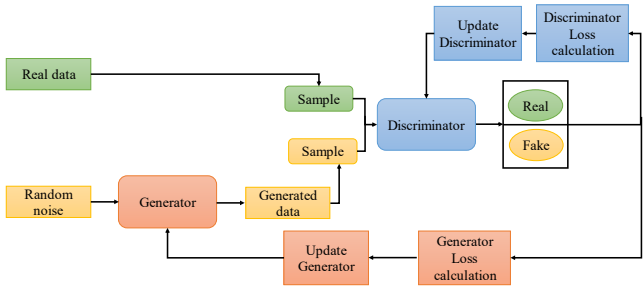

Fig. 10. Synthetic data generation using a generative adversarial network (GAN) model.

Once the training of the GAN model has ended, the generator is used to generate synthetic data. To assess the quality of the synthetic data, the evaluation metrics percent root mean square difference (PRD), root mean square error (RMSE) and the Fréchet distance (FD) are adopted. PRD is used to evaluate the difference between the real data and the generated data, and it can be formulated as:

$$PRD = \sqrt{\frac{\sum_{i=1}^{i=N}(x_i - \hat{x}_i)^2}{\sum_{i=1}^{i=N} x_i^2}} \times 100, \quad (13)$$

where $x_i$ denotes the value of the sampling point i of the real sequence, $\hat{x}_i$ represents the value of the sampling point i of the generated sequence, and $N$ is the length of the sequence.
RMSE quantifies the stability between the original data and the synthetic data, and it is expressed as:

$$RMSE = \sqrt{\frac{\sum_{i=1}^{i=N}(x_i - \hat{x}_i)^2}{N}} \quad (14)$$

FD measures the similarity between the real data and the generated data curves. Let $O_R = (u_1, u_2 \ldots u_R)$ be the order of points along the segmented real curves, and $O_S = (v_1, v_2 \ldots v_S)$ be the order of points along the segmented synthetic curves. The length $\| d \|$ of the sequence consisting of couple of points $\{(u_{a_1}, v_{b_1}), \ldots (u_{a_n}, v_{b_n})\}$ is computed as:

$$\| d \| = \max_{i=1,\ldots n} d(u_{a_i}, v_{b_i}), \quad (15)$$

where $d$ is the Euclidean distance.
FD is calculated as:

$$FD(R, S) = \min \{ \| d \| \} \quad (16)$$

Please note that a good or optimum synthetic data generation method should have very low or ideally close to zero $PRD$, $RMSE$, and $FD$ metrics.

Figure 11 shows that the different evaluation metrics are very small, which demonstrates that the synthetic data is very close and similar to the real data.

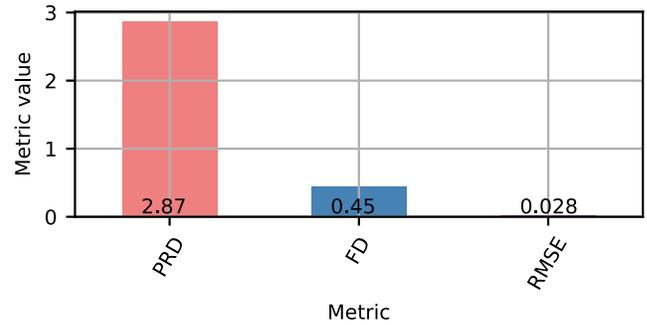

Fig. 11. Assessment of the synthetic data using the metrics PRD, FD and RMSE.

To qualitatively assess how close the distribution of the synthetic data is to the real data's distribution, the t-distributed stochastic neighbor embedding (t-SNE) [18], a technique for visualizing a high dimensional data into two-dimensional space (tSNE1 and tSNE2), is used. Figure 12 illustrates that the distribution of the synthetic data resembles that of the original experimental data, which proves the effectiveness of the generator in producing realistic data.

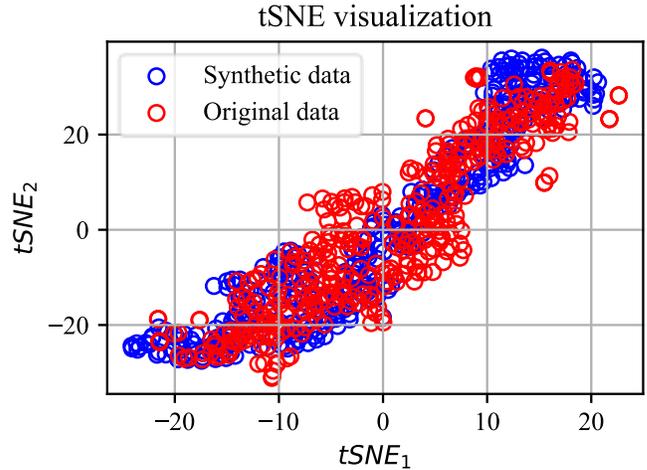

Fig. 12. t-SNE visualization of the synthetic and real data distributions

Figure 13 shows the histograms of the normalized current magnitudes of the synthetic and original data after 80 h. It can be seen that both histograms are close and that the synthetic current values are within the range of the real currents.

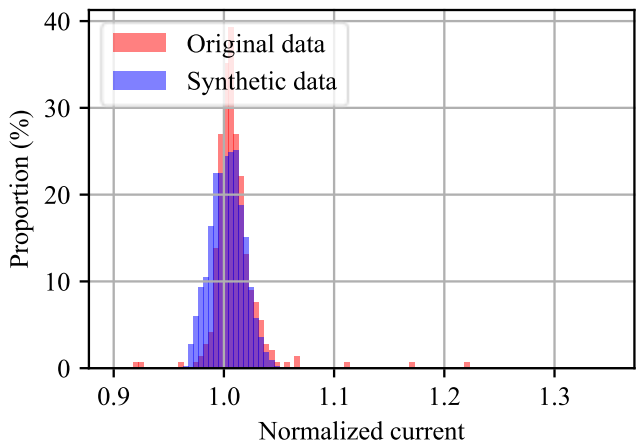

Fig. 13. Normalized current histograms of the real and synthetic data after 80 h.

In total, a synthetic dataset of 5,600 samples is generated. The RUL and the state of the device are computed for each sample based on the defined failure criteria. The said data is then normalized and fed to the ML models for training.

### D. Validation results of the performance prediction model

The proposed model is compared to other ML techniques, namely Multilayer perceptron (MLP), CNN, and RNN by adopting as evaluation metrics the mean absolute percentage error (MAPE) and the coefficient of variation of the root mean squared error (CVRMSE), which are formulated as follows:

$$MAPE = \frac{100}{N} \sum_{i=1}^{i=N} \left| \frac{x_i - x_i'}{x_i} \right| \quad (17)$$

$$CVRMSE = \frac{100}{\bar{x}} \sqrt{\frac{\sum_{i=1}^{N}(x_i - x_i')^2}{N}} \quad (18)$$

where $x_i'$ and $x_i$ denote the predicted and the true current values respectively. $N$ represents the number of test samples. $\bar{x}$ is the average of the true current values. It is to be noted that a lower value of MAPE and CVRMSE indicates a better prediction capability.

The different ML models are trained with the synthetic dataset and tested with the real experimental dataset. The results of the comparison illustrated in Fig. 14 show that the proposed model achieves the smallest values of MAPE and CVRMSE, which proves that the proposed method yields better prediction performance.

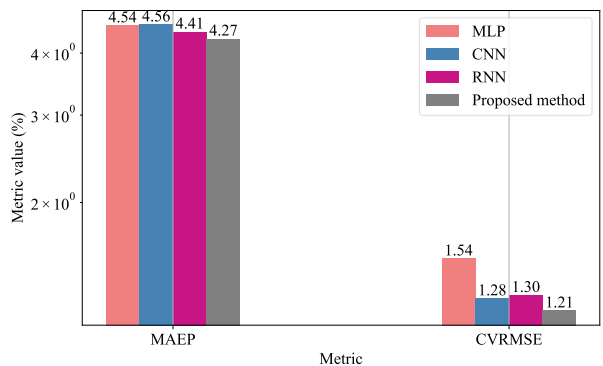

Fig. 14. Results of the comparison of the proposed method with MLP, CNN and RNN in terms of MAPE and CVRMSE.

The comparison of the results of computational inference time between the proposed model and the other methods are shown in Table I. As it can be seen, the proposed ML model consumes slightly more time than MLP and CNN models due to its deeper architecture, however, it executes faster than the RNN method.

TABLE I
COMPUTATIONAL TIME OF PROPOSED MODEL AND OTHER METHODS.

| Method | Inference time (270 samples) |
|---|---|
| MLP | 0.03 s |
| CNN | 0.08 s |
| RNN | 0.35 s |
| Proposed model | 0.28 s |

We evaluate the short-term and long-term prediction capability of the proposed model. Note that the model is trained with current measurements till 1000 h, and that it is tested to forecast the current values at 1500h, 2000h and 3000h. As shown in Fig. 15, the ML model accurately predicts the next value of the current measurement (one step prediction) with small prediction errors with a mean of 0.004 and a standard deviation of 0.027, and forecasts the next three values of the current measurements for the next time frames by achieving low prediction errors with a mean of -0.029 and a standard deviation of 0.032.

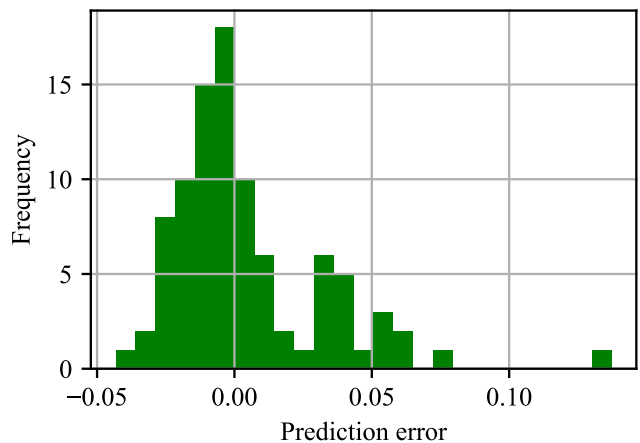

(a)

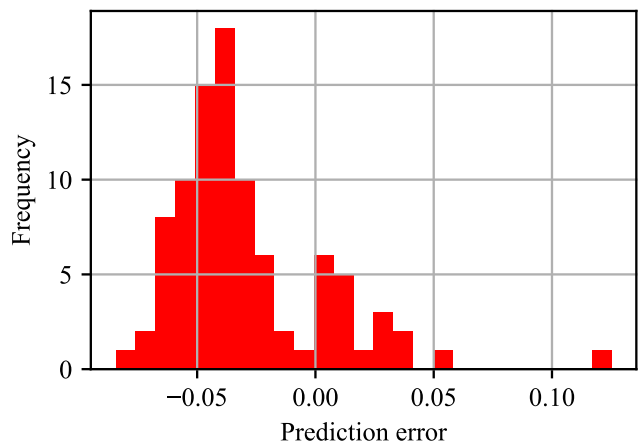

(b)

Fig. 15. Histograms of prediction errors: (a) for short-term prediction (one step forecasting), and (b) long-term prediction (multi-step forecasting).

Figure 16 shows the predicted values of two random samples. It can be observed that the forecasted values are close to the actual values and that they are following the same degradation trend as the actual values, which demonstrates the effectiveness of the proposed model in predicting the current measurements.

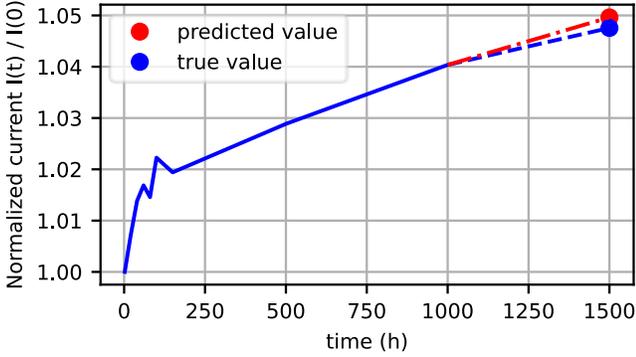

(a)

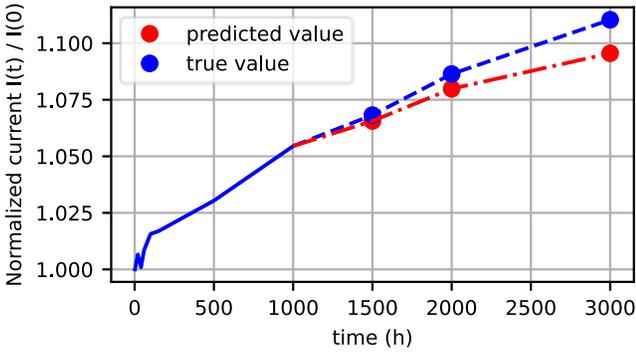

(b)

Fig. 16. Results of laser current prediction for two random test samples: (a) short-term prediction of the current at 1500 h, (b) long-term prediction of the current up to 3000 h.

Instead of adopting the one-step prediction ML model multiple times for performing long-term prediction, whereby the prediction for the prior time step is used as an input for making a prediction on the following time step, we investigated the capability of the ML model in performing a multi-step forecasting by predicting the entire forecast sequence in a one-shot manner. Figure 17 shows the adjusted architecture of the ML model for performing two step prediction. The two step ML model is trained as well with the synthetic dataset and tested with the experimental dataset. The test results show that the two-step model achieves higher values of CVRMSE (4.8%) and MAPE (5.8%) compared to the performance yielded by the one step model, which proves that it is less accurate.

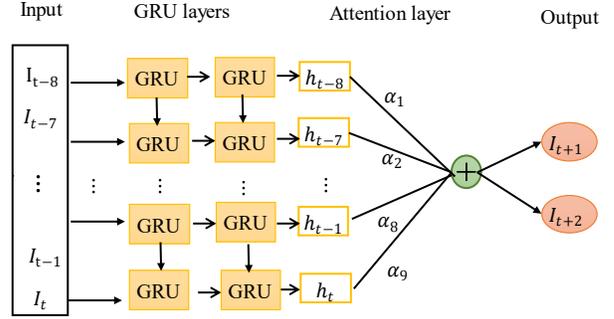

Fig. 17. Architecture of the two-step ML model for performance degradation prediction.

We investigated the impact of the input sequence length on the performance of the one step ML model. We trained the ML model with sequences of length 5, 6, 7, and 8, respectively. Figure 18 shows that the ML model's performance in terms of CVRMSE decreases with reducing the input sequence length. Reducing the input sequence length leads to a loss of the information representing the degradation trend, which impacts the capability of the ML model in capturing the relevant features for accurate prediction.

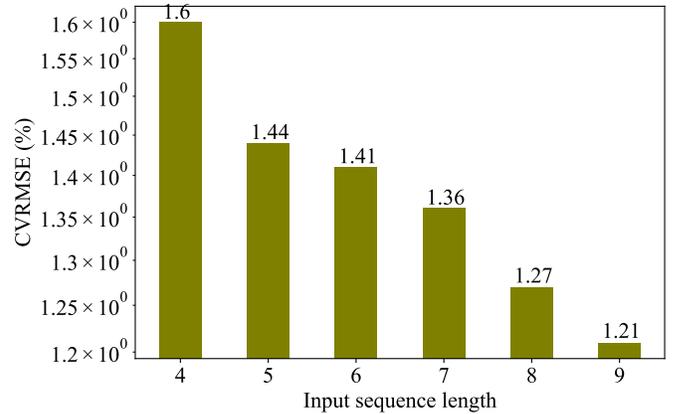

Fig. 18. Impact of the input sequence length on the performance of the ML model for performance degradation prediction.

### E. Validation results of the ML model for anomaly detection

The ML model for anomaly detection is trained with synthetic data, modelling the normal behavior of laser devices. After the training, the model is tested with experimental data of normal and anomalous lasers. To assess the anomaly detection capability, the following metrics are adopted:

- Precision (P) quantifies the relevance of the predictions made by the ML model. It is expressed as:

$$P = \frac{TP}{TP + FP}, \qquad (19)$$

where $TP$ denotes the number of "anomalous" sequences correctly classified, and $FP$ represents the number of "normal" sequences misclassified as "anomalous".

- Recall (R) provides the total relevant results correctly classified by the ML model. It is formulated as:

$$R = \frac{TP}{TP + FN}, \quad (20)$$

where $FN$ denotes Number of "anomalous" sequences misclassified as "normal".

- F1 score is the harmonic mean of the precision and recall, calculated as:

$$F1 = 2 \cdot \frac{P \cdot R}{P + R} \quad (21)$$

- The accuracy (A) can be defined as the total number of correctly classified instances divided by the total number of test instances. It is calculated as follows:

$$A = \frac{TP + TN}{TP + TN + FP + FN}, \quad (22)$$

where $TN$ denotes the number of "normal" sequences correctly classified.

The detection capability of the model is optimized by selecting the optimal threshold $\theta$. Figure 19 illustrates the precision, recall and F1 score (i.e. the harmonic mean of the precision and recall) curves along with $\theta$. It can be seen that there is a tradeoff between precision and recall. If the chosen threshold is higher than 0.025, many normal laser devices will be classified as anomalous devices, resulting in higher false negative and low recall scores. Whereas if the selected threshold is less than 0.015, many abnormal devices will be classified as normal, leading to higher false positive and a low precision score. Therefore, the optimal threshold, that provides the best precision and recall tradeoff (i.e. maximizing the F1 score) is selected. For the optimal chosen threshold of 0.019, the precision, the recall, the F1 score and the accuracy are 96.72%, 92%, 94% and 94.24%, respectively.

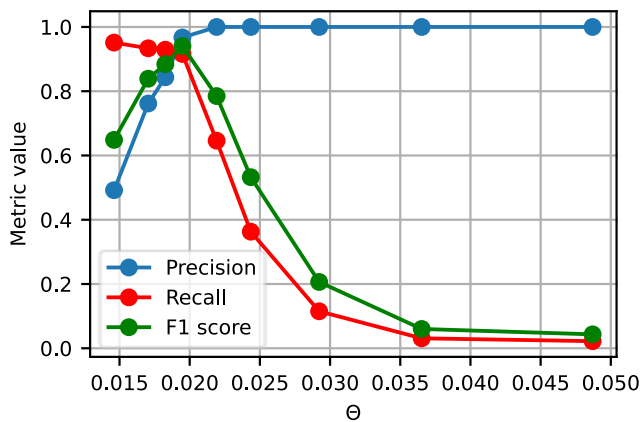

Fig. 19. The optimal threshold selection based on the precision, recall and F1 score scores yielded by the autoencoder.

The influence of reducing the input sequence length on the performance of the ML model for anomaly detection is investigated. Figure 20 illustrates that decreasing the input sequence length leads to a reduced F1 score of the autoencoder model. Reducing the sequence length too much (lower than 7) can cause the loss of the information underlying the normal behavior trend leading to underfitting and thereby worsening the performance.

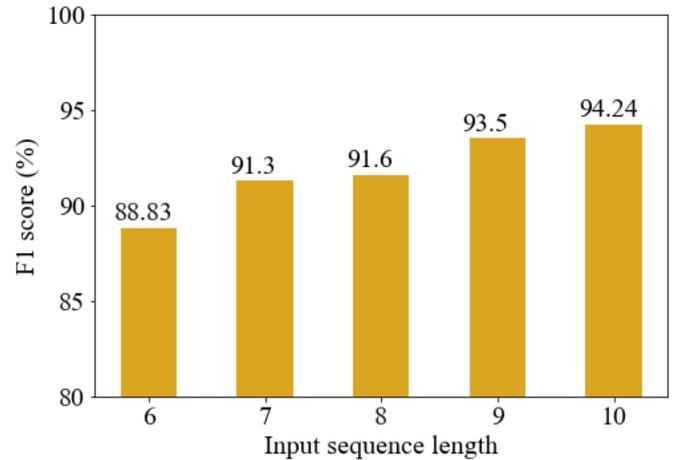

Fig. 20. Influence of the sequence length on the performance of the ML model for anomaly detection in terms of F1 score.

It is to be noted that the autoencoder model consumes 0.34s for performing the predictions given 247 test samples.

*F. Validation results of the ML model for RUL prediction*

After training the proposed model for RUL prediction using synthetic data is tested with experimental data by adopting the RMSE and the MAE evaluation metrics. We evaluate the prediction capability of the proposed method with the GRU model without attention mechanism and the attention based GRU method. Figure 21 shows that the proposed model achieves the lowest scores of RMSE and MAE, which proves adding the statistical features and the attention mechanism helps to enhance the RUL estimation capability. Adding the attention mechanism boosts the performance by achieving 10.3% and 8% improvements in RMSE and MAE metrics respectively. Whereas including both statistical features and attention mechanism enhances further the prediction capability by yielding 32.6% and 18.5% improvements in RMSE and MAE respectively.

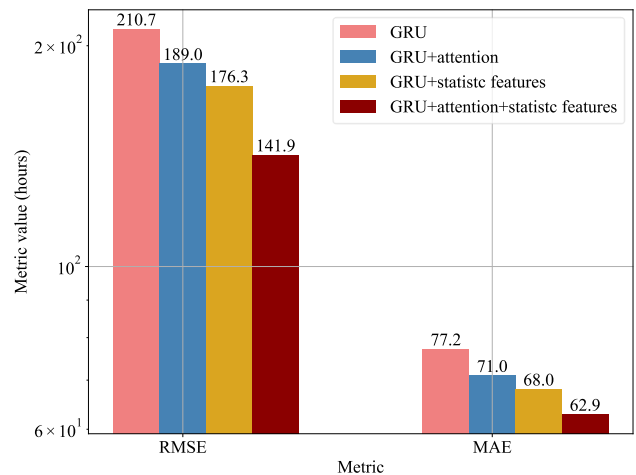

Fig. 21. Results of the comparison of the proposed model (GRU+ attention+ statistic features) with GRU model, attention based GRU method, and GRU + statistic features model using the RMSE and the MAE metrics.

The proposed model is compared with other ML techniques,

namely random forest (RF), support vector regression (SVR), MLP, RNN, LSTM and CNN, using as evaluation metrics the RMSE and the MAE. The results shown in Table II demonstrate that the proposed method outperforms the other ML models by achieving the smallest scores of RMSE and MAE. The comparison results of the computational time (inference time) of the proposed model and other ML methods in Table II show that the proposed method consumes more time in testing (inference) due to its deep architecture, and that the shallow ML techniques RF and SVR are much less time consuming in testing.

TABLE II
RESULTS OF THE COMPARISON OF THE PROPOSED METHOD WITH OTHER ML TECHNIQUES USING RMSE, MAE, AND COMPUTATIONAL TIME.

| Method | RMSE | MAE | Inference time (55 samples) |
| --- | --- | --- | --- |
| RF | 1269.37 | 882.67 | 0.003 s |
| SVR | 973.8 | 547.77 | 0.004 s |
| MLP | 309.38 | 120.57 | 0.035 s |
| RNN | 228.12 | 70.25 | 0.71 s |
| CNN | 253.26 | 95.4 | 0.148 s |
| LSTM | 181.33 | 93.68 | 0.49 s |
| **Proposed method** | **141.9** | **62.85** | **0.75 s** |

The proposed approach achieves also better prediction capability compared to the recently presented CNN-LSTM model for laser RUL estimation [12] (RMSE = 385 hours, MAE = 261 hours) by providing 63.14 % and 75.9% improvements in RMSE and MAE metrics respectively.

To assess further the RUL estimation capability of the proposed model, we compare the predicted RUL values to the true RULs at different stages of degradation. As shown in Fig. 22, the RUL values estimated by the model are very close to the true RUL values, which proves the effectiveness of the proposed model in accurately predicting the RUL of the laser device.

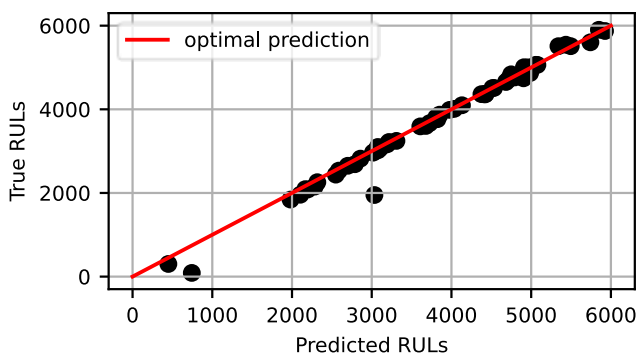

Fig. 22. Results of Predicted RULs by the proposed model vs. actual RULs.

The impact of reducing the length of the input sequence on the performance of the ML model for RUL prediction is explored as well. Figure 23 shows that the prediction capability of the ML model decreases with the reduction of the length of the input sequence due to the loss of the relevant information describing the degradation trend which adversely impacts the performance.

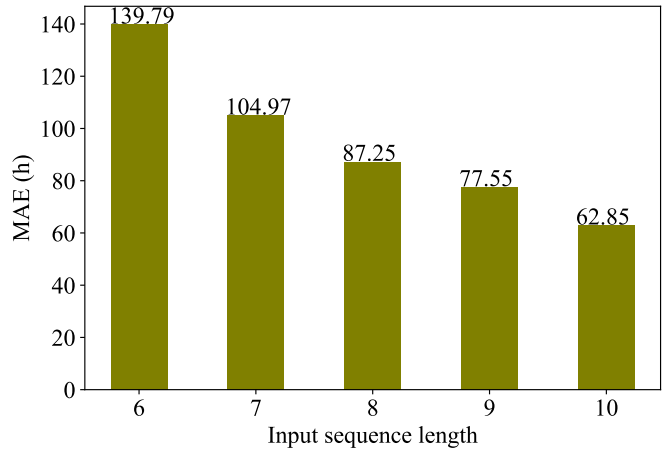

Fig. 23. Influence of the sequence length on the performance of the ML model for RUL prediction.

V. CONCLUSION

An ML-based predictive maintenance framework for semiconductor lasers is proposed for real-time monitoring and prognosis of the laser device during operation. The proposed approach contains three main steps: real-time performance degradation prediction, degradation detection and RUL prediction. An attention based GRU model is used to predict the laser performance degradation (i.e., the laser current increase). The convolutional autoencoder is adopted to detect any degradation or abnormal behavior of the laser. An attention-based deep learning model is used to estimate the RUL of the laser. The different models are trained with synthetic data generated by the GAN model, and tested with experimental data of tunable lasers. The results demonstrate that the attention-based GRU model achieves a good degradation performance prediction (RMSE of 0.01), the convolutional autoencoder yields a high detection accuracy of 94.2%, and the attention-based deep learning model achieves a good RUL estimation (RMSE of 142 hours), which demonstrates the effectiveness of the proposed framework. The results show also that adding statistical features underlying the degradation trend helps to improve the performance of the RUL prediction model, and that adding the attention mechanism enhances the prediction capability. The results demonstrate as well that the GAN is able to produce laser reliability data that is close to the real experimental data, and in case of limited in-field data or experimental data, synthetic data generated by GAN is a good solution to train the ML model, and that the performance of the ML model trained with the synthetic data is good when tested with real data. The same concept of the proposed framework is readily applicable to other optoelectronic devices such as semiconductor optical amplifiers due to the similarity of their structures.